\documentclass{article}

\usepackage{arxiv}

\usepackage[utf8]{inputenc} 
\usepackage[T1]{fontenc}    
\usepackage{hyperref}       
\usepackage{url}            
\usepackage{booktabs}       
\usepackage{amsfonts}       
\usepackage{amsmath}
\usepackage{nicefrac}       
\usepackage{microtype}      
\usepackage{lipsum}		
\usepackage{graphicx}
\usepackage{doi}

\newcommand{\supplementarysection}{%
  \setcounter{figure}{0}
  \setcounter{section}{0} 

  \let\oldthefigure\thefigure
  \renewcommand{\thefigure}{S\oldthefigure}
  \section{Appendices}
}

\title{An Artificial Neural Network Functionalized by Evolution}

\author{\href{https://orcid.org/0000-0003-1997-1141}{\includegraphics[width=0.025\linewidth]{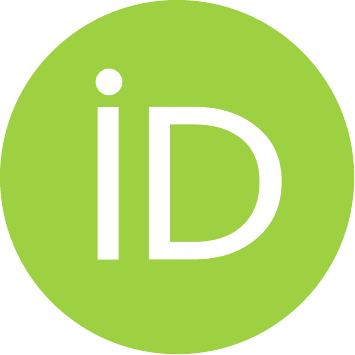}\hspace{1mm}Fabien Furfaro}\thanks{Code is available on Github at the following repository: github.com/fabienfrfr/AFF; Python module named FUNCTIONNAL\_FILLET.}\\
	Independent scientist\\
	\texttt{fabien.furfaro@gmail.com}\\
	\And
	\href{https://orcid.org/0000-0002-4449-8117}{\includegraphics[width=0.025\linewidth]{orcid.pdf}\hspace{1mm}Avner Bar-Hen}\\
	CEDRIC\\
	CNAM\\
	75003 Paris, France\\
	\texttt{avner@cnam.fr} \\
	\And
    \href{https://orcid.org/0000-0003-4036-6114}{\includegraphics[width=0.025\linewidth]{orcid.pdf}\hspace{1mm}Geoffroy Berthelot}\\
	INSEP, RELAIS\\
	75012 Paris, France\\
	\texttt{geoffroy.berthelot@insep.fr}\\
}

\date{}

\renewcommand{\shorttitle}{An ANN Functionalized by Evolution}

\hypersetup{
pdftitle={An ANN Functionalized by Evolution (preprint)},
pdfsubject={cs.AI},
pdfauthor={Fabien Furfaro, Avner Bar-Hen, Geoffroy Berthelot},
pdfkeywords={ANN, NAS, Evolution, MNIST, Markov Decision Process},
}

\begin{document}
\maketitle

\begin{abstract}
The topology of artificial neural networks has a significant effect on their performance. Characterizing efficient topology is a field of promising research in Artificial Intelligence. However, it is not a trivial task and it is mainly experimented on through convolutional neural networks. We propose a hybrid model which combines the tensor calculus of feed-forward neural networks with Pseudo-Darwinian mechanisms. This allows for finding topologies that are well adapted for elaboration of strategies, control problems or pattern recognition tasks. In particular, the model can provide adapted topologies at early evolutionary stages, and 'structural convergence', which can found applications in robotics, big-data and artificial life.
\end{abstract}


\section{Introduction}
With the recent and continuous sophistication of AI methods \cite{stanley2019designing} and the increase in computing power, the question of Life and the emergence of general intelligence developed \cite{lehman2020surprising}. The question of machine learning with neural networks was studied during the 1950s, with the introduction of Hebbs\textquoteright rules. These are the essence of modern AI, which now features sophisticated simulation methods \cite{fan2019brief}, the universal approximation theorem \cite{hornik1989multilayer} and gradient descent \cite{amari1993backpropagation}, among others. These methods have demonstrated AI's capabilities in solving many different problems \cite{siegelmann1994analog}. However, these methods generally optimize the learning independently of the topology. In other words, they do not optimize the structure of the layers. Here, we draw inspiration from the efficiency of the universal approximation theorem \cite{hornik1989multilayer} and brain module \cite{rakison2008infants, ohman2001fears} to characterize the minimal rules which favors the emergence of a functional brain-like structure. To date, two hypotheses are dominant: cognitivism where thinking is analogous to an information processing process of neural structure \cite{sep-computational-mind}; and connectionism where mental phenomena are emerging processes of networks of single interconnected units \cite{thagard1996cognitive}. Based on these two hypotheses, we use pseudo-Darwinian mechanisms to design 'functional' ANN with the aim of reproducing an evolutionary system which allows for the emergence of an adapted strategy. Functional means the ANN is shaped for a specific task throughout evolution. More precisely, the topology of an artificial neural network is evolved for designing a network structure that provides good performance, where a performance is typically a score. This results in the network self-adapting its topology for a specific task. Such topological optimization, or NAS, is a field of active and promising research \cite{stanley2019designing}, with approaches like Neuroevolution of Augmenting Topologies (NEAT) \cite{stanley2002evolving}, Graph Neural Networks (GNN) \cite{zhou2020graph, asif2021graph} and Automated Machine Learning (AutoML) \cite{he2021automl}.

Here we introduce a new methodology in the AutoML framework, where (\textit{i}) the connections between layers of neurons are evolved and connections can be permuted, similarly to evolutionary mechanisms; (\textit{ii}) we take advantage of the multi-layer perceptron methods, allowing for large input-outputs layers, tensor calculation and gradient descent for training the weights. In some other evolutionary approaches, weights can be evolved along with the topology or separately using the crossover algorithm, which can produce very specialized networks \cite{stanley2002evolving}. In this work, the weights are trained between each new generation, providing a middle ground between pure evolutive algorithms (which correspond to a single training cycle per generation) and gradient descent (which correspond to one generation or no evolution). And (\textit{iii}) we do not use convolutional neural networks and instead rely on typical feed-forward artificial neural networks with memory (using a recurrence link), which can be used in various tasks with different levels of complexity: pattern recognition, signal processing, non linear time dynamics systems, etc. \cite{sengupta2020review}. The performance of the proposed approach is measured through 'parents', which are evolutionary networks maintained throughout evolution because they present an adaptive advantage. These are compared to 'control' networks, which are typical and non-evolving ANNs. In particular, we evaluate the performance of this approach in 3 tasks: a numerical game (or TAG-game) involving prey-predator dynamics, which illustrates the effects of pseudo-Darwinian evolution for natural adaptation \cite{reynolds1994competition}; a control problem, which is the typical cart-pole problem as described by Barto, Sutton, and Anderson \cite{barto1983neuronlike}; and a pattern recognition task using the MNIST database, which is a standard problem for studying the effectiveness of deep learning models \cite{deng2012mnist}. Both the TAG-game and cart-pole tasks are Markov Decision processes, whereas the MNIST problem is a typical pattern recognition and classification task. The capability to perform different tasks is measured through the TAG-game, by studying the adaptation to either be the predator (alt. the prey) or both (\textit{i.e.} alternating between prey and predator). Finally, the evolutionary convergence in terms of topology is assessed using the absolute number of descendants in the corresponding phylogenetic tree.

\section{Material and Methods}
\begin{figure}
\centering
\includegraphics{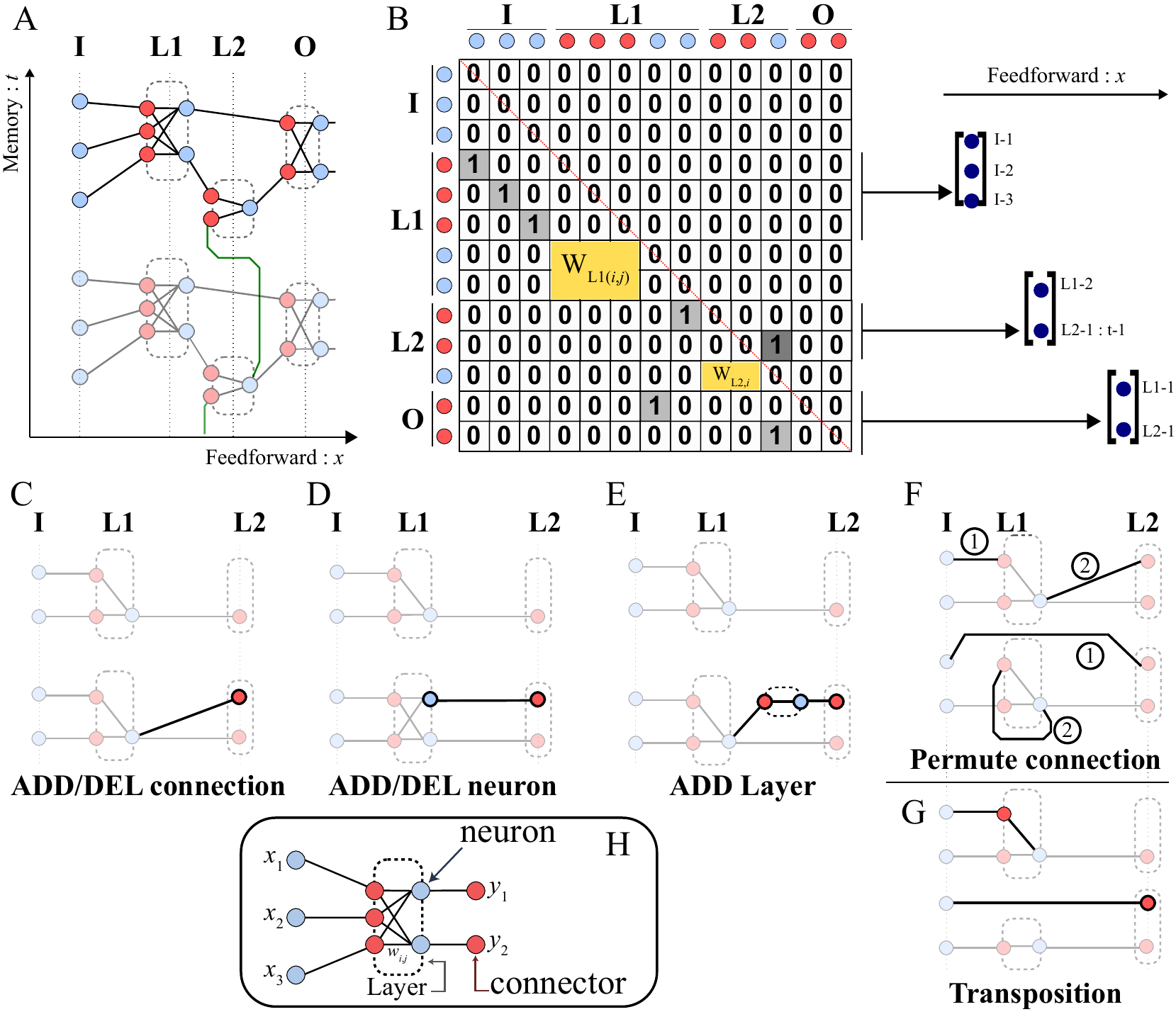}
\caption{A directed graph is randomly constructed in \textbf{A}, and the corresponding adjacency matrix is given in \textbf{B}. Red nodes correspond to inputs of each layers and blue nodes are the neurons (perceptron nodes) (output of a layer). The green link is the recurrence or virtual link. Capital letters \textbf{I}, \textbf{L} and \textbf{O} correspond to (I)nput, (O)utput and (L)ayer respectively. Mutations are illustrated in panels \textbf{C}-\textbf{G}. In each of these panels, the highlighted elements correspond to the mutated elements. An example is provided in panel \textbf{H}, which explains the significance of each symbol.}
\label{fig:2}
\end{figure}

\subsection{Construction of the network}
The first step is to construct the topology of the network that will be evolved. We start by a single, directed graph that meet certain criteria to allow for applying ANN methods such as gradient descent, retroprogation, and other (See Appendices). The topology of this graph is described by an adjacency matrix that contains the links between the different layers, a recurrence link \cite{werbos1988generalization} and the biases (Fig. \ref{fig:2}). From this adjacency matrix, a network can be constructed and 'functionalized' through evolution.

\subsection{Functionalization using Pseudo-Darwinian Evolution}
\begin{figure}
\centering
\includegraphics{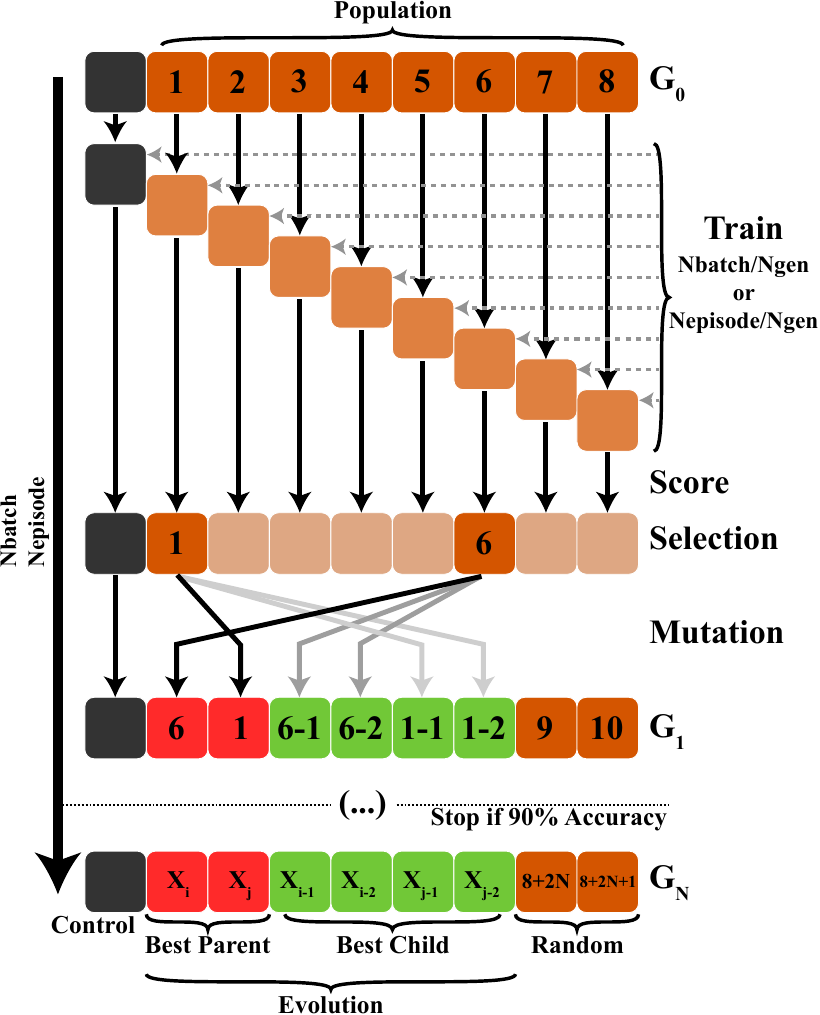}
\caption{Illustration of the algorithm. An initial population of $n=9$ networks, with 8 networks and 1 control network is generated (see 'Pseudo-Darwinian Evolution' section, step 1) and trained (step 2). The next step is to evaluate the two best networks given their performance (score, loss function, etc.) (step 3). These two networks become 'parents' of a new population of 4 children (step 4) and two new random networks are introduced (step 5). This population becomes the new generation, here labeled $G_1$. The process is repeated for $N$ generations, until a performance threshold is met.}
\label{fig:4}
\end{figure}

We aim at reproducing the evolutionary mechanisms that gave rise to the functional structure of the brain. Our approach rely on the methods initially developed for ANN (back-propagation, gradient descent, etc.) but we add an evolutionary process for selecting the connections between the different layers. This allows for the discovery of new zoology of neuron network structures while keeping the advantages of tensor calculus. The general algorithm for constructing an evolutionary functional network is illustrated in Fig. \ref{fig:4} and can be summarized as follow:
\begin{enumerate}
    \item draw $n$ random realizations of neural networks, from which $\lfloor \sqrt[4]{n} \rfloor$ will serve as control networks. This correspond to the initial population networks at the first generation ($G_0$)
    \item train the current population using gradient descent and a number of batches
    \item select the $\alpha$ most efficient networks. The algorithm stops if either of the following two conditions is met: (1) the accuracy/score is 'good enough'; (2) the predefined maximal number of generations $N$ is reached. This threshold number is defined by the number of batches required for the control networks for solving the same problem. Otherwise we continue.
    \item create $\alpha^2$ independent children, inheriting the structure (traits) from the selected parents (plasticity steps) networks and mutate their structures (Fig \ref{fig:2}, \textbf{C}-\textbf{G})
    \item insert $\alpha$ new random networks
    \item return to step (2) with this new population of $n$ networks and start a new generation.
\end{enumerate}
'Good enough' means that the network with the highest performance is matching a pre-defined performance criterion. The parameter $\alpha$ is set to $\alpha = \sqrt{n} -1$ for simplicity. The number of control networks is arbitrary set to $\lfloor \sqrt[4]{n} \rfloor$. The mutations in step 4 are defined in section \ref{section_mutations}.

\subsection{Mutations}
\label{section_mutations}
Each mutation is required to meet two conditions: first, the mutation cannot creates 'dead ends', meaning that there should always exist a path between any given pair of nodes in the network. In other words, the minimal degree of each node is 2, to the exception of the main input and output nodes of the network. Second, the initial structure cannot be altered. By initial we meant the neural network used in the first generation (see step 1 of the list in the previous section). Seven types of mutations are defined: (\textit{i}) add a single connection to an existing neuron (Fig. \ref{fig:2} C); (\textit{ii}) Add a neuron to one of the existing layers (Fig. \ref{fig:2} D); (\textit{iii}) Add an additional layer (Fig. \ref{fig:2} E). If the neuron of this layer connects backwards in time, it becomes equivalent to a 'virtual' input generator (see green link in Fig. \ref{fig:2} \textbf{A} and Appendices). (\textit{iv}) Remove a 'duplicate' connection, that is a connection that already exists between two nodes. (\textit{v}) Remove a neuron. (\textit{vi}) Permutation of existing connections (\textit{i.e.} swapping two connections) (Fig. \ref{fig:2} F). (\textit{vii}) Move or transpose an existing connection (Fig. \ref{fig:2} G).

\subsection{TAG-Game, cart-pole and MNIST}
A simple prey-predator TAG-Game is implemented such that a network or 'agent' can evolve and learn efficient strategies for evading the predator and/or catching the prey. The aim is to reproduce survival strategies which are observed in living organisms, in particular the strategies of predation and anti-predation. This practically results in performing all the methodological steps described in the previous sections. This TAG-Game takes place in an environment that is a two-dimensional grid or lattice of size $N\times M$ cells with periodic boundary conditions. The agent is trained to play both the predator and prey role against a simple, non-learning, algorithm that always minimizes or maximizes the Euclidean distance between itself and the agent. Each 'life-cycle', the agent alternates roles, where a 'life-cycle' corresponds to a complete restart of both the position and the state. We also study specialization, where the agent either learns to be a predator (predator specialization) or a prey (prey specialization). The game rules and rewards are detailed in Appendices.

The cart-pole task is a typical benchmark problem for control purposes, previously described by Barto, Sutton and Anderson \cite{barto1983neuronlike}. The aim of this task is to balance a pole that is hinged to a movable cart by applying forces to the cart's base. The cart moves along a friction-less track and the system is controlled by applying a force of +1 or -1 to the cart. The pendulum starts upright, and a reward of +1 is provided for every time-step that the pole remains upright. The episode ends when the pole is more than 15 degrees from vertical, or the cart moves more than 2.4 units from the center.

For theses two Markov decision processes, we fix the exploration/exploitation dilemma using a custom routine (see Appendices).

Finally, The MNIST database is used for pattern recognition. The same approach is used, except we only use one control network, due to limitations in computational resources.

\subsection{Training Dataset}
Network training is performed as follow for the three tasks: (TAG-game) The total experiment consists in $12800$ episodes, which are divided in 100 generations, yielding 128 episodes per generation. The number of time steps is 256, divided by 16 life-cycles. The batch size is 16 time steps. (Cart-pole) The total experiment consists in $25000$ episodes, divided in 100 generations thus yielding 250 episodes per generation. The number of batches by episode is defined by the number of time steps where the pole is considered upright. The maximum number of time steps is 300, and reaching this value corresponds to a successful try. The batch size is 25 time steps, and the maximum number of batches per episode is 12. (MNIST database with 60,000 samples), we repeated the training of the base only for 2 epochs, split into 100 generations of batch 50, or 2400 batches in all, with 24 batches per generation. The initial size of the population is 49 for both the TAG-game 49 and the pole-cart, and $\{25-49\}$ for the MNIST database.

\subsection{Convergence}
In particular, two types of convergences are studied: (\textit{i}) the convergence of the model towards a functionalizable problem in terms of loss and score and (\textit{ii}), the convergence of structures or adaptive strategies. For the second one, it is necessary to run several experiments in parallel and compare them to a control ANN -a typical fully interconnected single-layer linear network- and a random network generated at each cycle. Two indicators are introduced for describing the convergence of the model: the loss function for the classification data and the score for the cart-pole and TAG-game, defined by the proportion of successes per life-cycle. The lineages of each network is studied through a phylogenetic tree which is constructed from the networks in all generations. In particular, we measure the number of descendants $\#\text{nodes}$ in each lineage $L$ using:
\begin{equation}
\label{eq:eq1_Lineages}
    \#\text{nodes}_L = \dfrac{1}{2} \sum_{i \in L} k_i
\end{equation}
with $k_i$ the degree of node $i$ (Fig. S1, also see Appendices). We study the sum of the lineages of the parents, children and both (Fig. \ref{fig:3}).

\section{Results}
\begin{figure}
\centering
\includegraphics{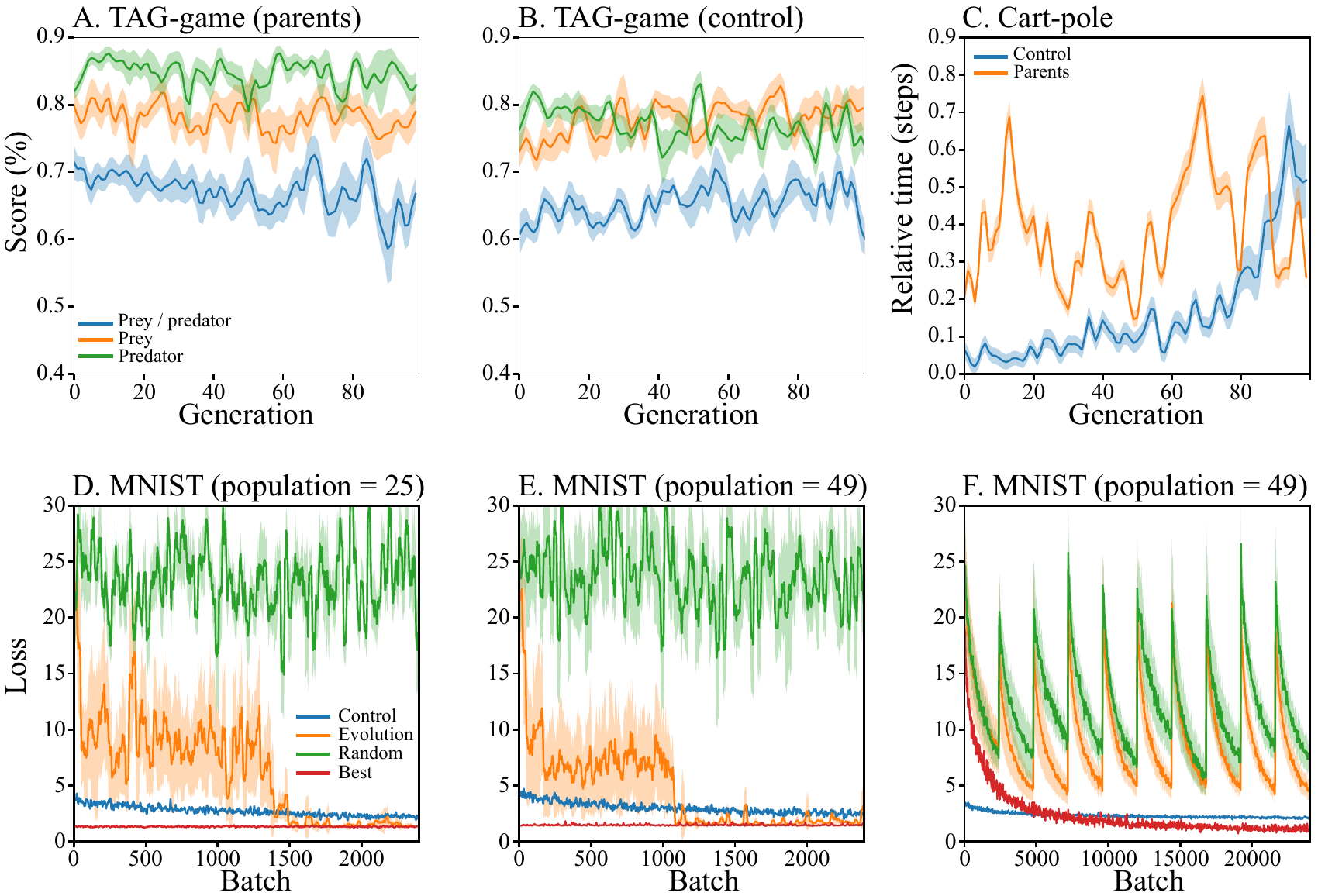}
\caption{Average performance of the proposed approach for the TAG-game (panels \textbf{A}-\textbf{B}), the cart-pole \textbf{B} and the pattern recognition task using the MNIST database (panels \textbf{D}-\textbf{F}). The panel \textbf{A} represents the average score of the best parents at each generation when learning both the prey and predator role (blue line) or either the prey (orange) or predator role (green). The panel \textbf{B} represents the average score of the control networks, which are typical feed-forward ANNs. In panel \textbf{C} the average relative time (the number of time steps between successive tests) is given for the control (blue line) and the best parents (orange line). In panels \textbf{D}-\textbf{F}, the x-axis corresponds to the number of batches received by the network for training. The evolution corresponding to the best parents and best children (orange line). The random network (green line) is a network, randomly constructed at each generation (Fig. \ref{fig:4}). While both the random network and parents needs to re-train at each generation, the control benefits from the training of the previous generations (Fig. \ref{fig:4}). The best network (red line) is the network with the overall best performance, which corresponds to the performance of the best parents ever recorded. It is obtained after several generations and allows for a direct comparison with the control network (blue line). The initial population is 25 for panel \textbf{D} and 49 for panels \textbf{E}-\textbf{F}. There are 100 generations in panels \textbf{A}-\textbf{E}, whereas there are only 10 generations in panel \textbf{F}. The confidence interval corresponds to the standard deviation of the corresponding population.}
\label{fig:1}
\end{figure}

\subsection{TAG-Game performance and functionalization}
The dynamics of both the control and the parents networks shows relative overall stability for the three tasks across 100 generations (Fig. \ref{fig:1} \textbf{A}-\textbf{B}). The results are similar when learning the prey/predator or prey tasks for the parents and the control networks, with average score values around 65-70\% and 75-80\% success respectively. However, the parents show a better average performance compared with the control when learning the predator role, with values in the 80-85\% range for the parents and 70-80\% for the control networks. There is also a difference when learning the role: it appears that learning only one role -either be the prey or predator- is easier than learning both roles at once. In other words, it suggests that specialization is more efficient than learning multiple tasks at once \cite{zhang2018overview}. While such an observation is clear for the parents, it is less obvious for the control networks, where the average scores of the prey or predator roles overlap (Fig. \ref{fig:1} \textbf{B}). This may suggest that the evolutionary approach better accounts for the TAG-game specificities attached to a particular role, or better accounts for the game mechanisms.
\subsection{Cart-pole problem}
The evolutionary networks seem unstable as the average performance of the best parents shows high heterogeneity. In the opposite, the control networks learn in a smoother way (Fig \ref{fig:1} \textbf{C}). Interestingly, the evolutionary networks provide early solutions with the first peak occurring at the 10\textsuperscript{th} generation, whereas the control networks are reaching a solution at the end of the experiment, at the 90\textsuperscript{th} generation.
\subsection{Pattern recognition using MNIST}
The loss function of both the control and evolutionary networks (parents) decreases with the number of batches (Fig. \ref{fig:1} \textbf{D}-\textbf{E}). The single control network presents a smooth decrease whereas the evolutionary networks show a brutal collapse after some instability. When this 'structural convergence' occurs, the evolutionary networks become more efficient than both the control and the random ones. The structural convergence occurs faster when we have a population of 49 networks (1100 batches) than when we have a population of 25 networks (1500 batches). When increasing the number of batches while decreasing the number of generations to 10, we observe multiple peaks followed by exponential-like decreases in the loss function. This is due to the re-learning process occurring every 10 generations. (Fig. \ref{fig:1} \textbf{F}). Again, we notice that the best network decreases more efficiently than the control one. When training both the control and best network for 200 epochs, we obtain the following results: $89.4\% \pm0.71$ (control) vs. $93.2\% \pm1.37$ (best network).

\begin{figure}
\centering
\includegraphics{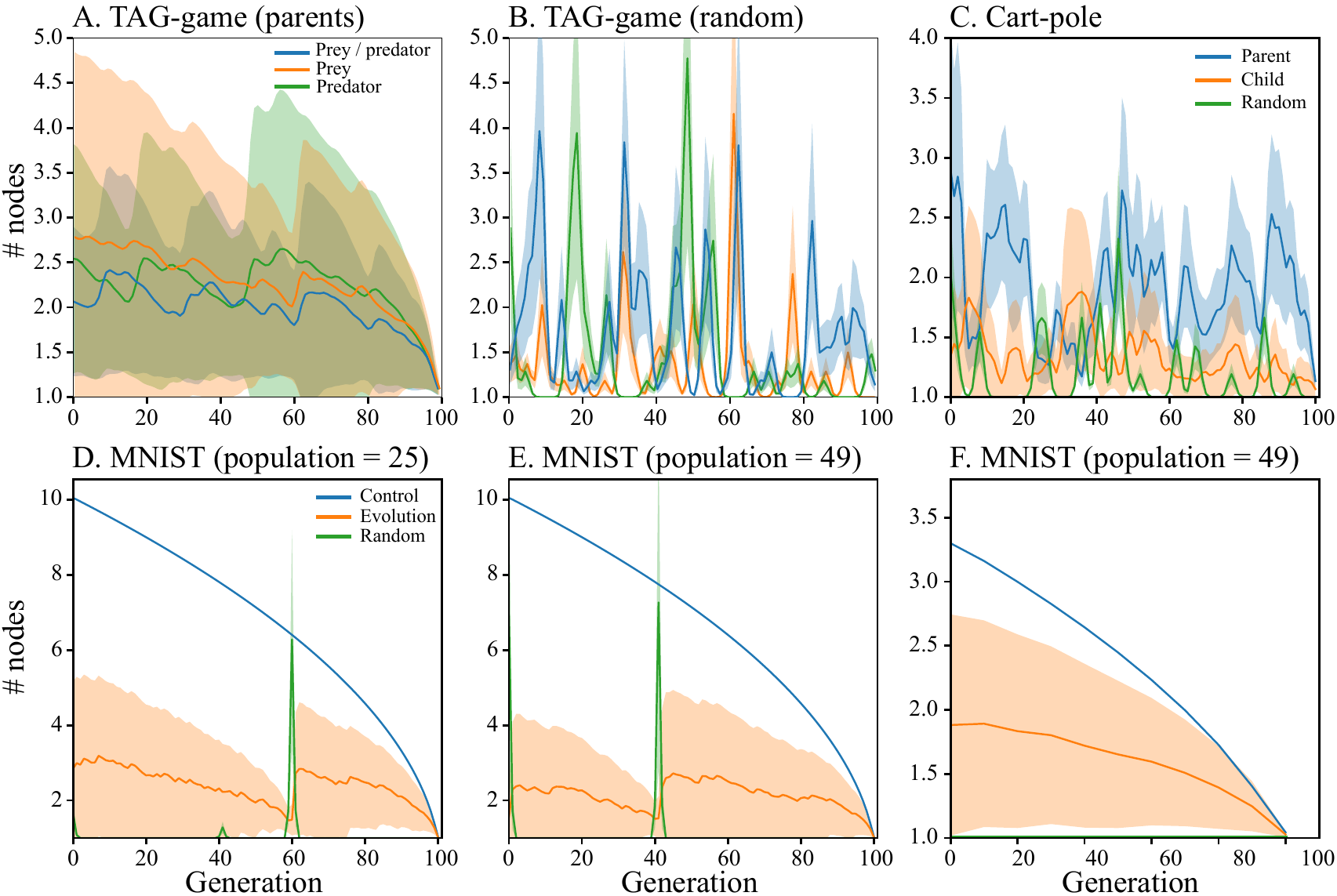}
\caption{Total number of descendants in the phylogenetic tree for the TAG-game (\textbf{A}-\textbf{B}), cart-pole (\textbf{C}) and MNIST (\textbf{D}-\textbf{F}). This allows for studying the dominance of efficient lineages. In panel (\textbf{A}) the average number of descendants is given for one experience and for the evolved networks only, thus excluding the control and random networks. In panel (\textbf{B}) the average number of descendants is given for the population of random networks only. It shows the dominant lineages originating from random networks. The average number of descendants for the cart-pole is given for the parents, children and random networks in order to see if some mutations provide a competitive advantage (\textbf{C}). The dominant (parents) lineage networks obtained in the generations 10-20 correspond to the efficient networks obtained in Fig. \ref{fig:1} \textbf{C}. In panels (\textbf{D})-(\textbf{F}), the orange line corresponds to the value for both the best parents and children.}
\label{fig:3}
\end{figure}

\subsection{Stability and convergence of selection model}
The stability of the evolutionary network is driven by the number of nodes in the network. For instance, a number of 784 inputs is used for the MNIST task, whereas only 4 and 17 inputs are used for the Cart Pole and TAG-game respectively (see Appendices: graph construction). This suggests that a higher number of nodes and links in the initial network can lead to evolutionary stability, in other words that lineages are less volatile.

The number of descendants in the phylogenetic tree allows for measuring both lineage duration and dominance. In other words, it measures for how many generations a lineage exists throughout evolution before being dominated by a more competitive lineage. It also measures how it spreads in the population, in terms of the absolute number of descendants at each generation, or evolutionary step. For the TAG-game, we observe a heterogeneity of the parent nodes, whatever the strategy (Fig \ref{fig:3} \textbf{A}). The random networks are shown to interfere in lineages quite frequently and for each strategy (Fig \ref{fig:3} \textbf{B}). It means that the evolution process is not able to select nor design a structure that remains dominant in terms of performance. According to the aforementioned stability hypothesis, using both larger networks -in terms of number of nodes- and an increased population size -in terms of the number of networks- could possibly help reducing the interference from random networks. However, it is not clear whether it is desirable or not, as random networks can also bring additional competitive topologies, which differs from the initial population. A similar result is obtained in the cart-pole task, where the random networks interfere in the evolutionary process (Fig \ref{fig:3} \textbf{C}). The random networks can produce a dominant linage, as shown in the pattern recognition task (Fig \ref{fig:3} \textbf{D}-\textbf{E}). This occurs in the two experiments with a different initial population size at generations 62 and 43 respectively. When the number of mutations is reduced to 10 (\textit{i.e.} 10 generations), the best network is obtained at the initial generation (Fig \ref{fig:3} \textbf{F}). However, this is a mechanistic effect as both the number of random networks at each generation and the probability to create a new dominant lineage depends on the population size and the number of generations.

\section{Discussion}
Four different results rise from this study: (\textit{i}) the larger the network, the more stable it is; (\textit{ii}) the greater the number of individuals in the initial population, the faster it converges; (\textit{iii}), a greater efficiency when separating roles in the TAG-game, suggesting that specialization is more effective; and (\textit{iv}), two types of convergence can be obtained: the first one is the 'structural convergence' occurring when increasing the number of mutations (or generations). The second one can be obtained by gradient descent when there is less mutations and more batches (Fig. \ref{fig:1} \textbf{D}-\textbf{F}). These two types of convergence can be optimized jointly by increasing the number of mutations and the number of batches at the same time. This results in a hybrid model, where optimization occurs in the two directions at once: \textit{i.e.} using gradient descent and structural optimization.

We showed that it is possible to apply such a hybrid model to pattern recognition and classification problems (MNIST). The efficiency of the best network is about $93.2\%$, whereas current top methodologies, based on convolutional neural networks, can reach an efficiency as high as $99.8\%$ \cite{hu2018squeeze}. However, we tested the approach on a single laptop with very limited computational resources. Using more computational power would allow for increasing the efficiency of our model, by increasing the initial population of networks and their size, and using a larger number of generations for instance. As our model includes recurrence, it would also be possible to use it for a range of classification problems such as language recognition, where it has already been shown that recurrence allows better convergence \cite{amberkar2018speech, hochreiter1997long}. However, in our approach, the construction of the graph is based on a random neuronal structure. It would be interesting to improve the construction of such a structure by using a process similar to what we observe in evolution \cite{ghysen2003origin} and development biology. In the development of any bilaterian organism -such as \textit{Drosophila} \cite{srahna2006signaling}, zebrafish \cite{kimmel1993patterning} or mice (\textit{Mus domesticus}) \cite{balaskas2012gene}- we have spatial patterns which are guided by the interaction between morphogens and gene regulatory network (GRN), that can be modeled using reaction-diffusion processes \cite{lefevre2010reaction}.

The TAG-game is an interesting task for evaluating the ability of the model to learn the prey/predator role or either the prey or predator ones. However, it is limited in both its design and the very simple mechanisms it uses. It is based on a 2 dimensional lattice with periodic boundary conditions and such boundaries conditions are not realistic in living systems. For example, borders can be useful for insects such as the American cockroach \textit{Periplaneta americana}, which are reported to follows surfaces such as walls \cite{cowan2006task}. Our model also allows for the appearance of a vestigial structure which is specific to living organisms, while not being the optimal structure for the gradient descent \cite{smith1985developmental}. For instance, in human, the recurrent laryngeal nerve passing by heart is not optimal. It is inherited from fish, and such an evolutionary organic structure is very common in the living \cite{jacob1977evolution}. Finally, some winning strategies are very efficient but not very elaborated: they simply exploit the flaws in the rules of the game. This is very frequent in reinforcement learning \cite{lehman2020surprising}. One way to improve the strategies elaborated by the model is to use a biological point of view, allowing for the emergence of a larger panel of strategies. A more realistic version of the TAG-game would take place in the same ecosystem with different trophisms, with more general rules of competition / cooperation. One can imagine a system where there are agents with different trophism / nutrient values (competition) and other agents with the same trophism and a probability of sharing resources when it feeds on lower trophism (cooperation). An example is a multicellularity model \cite{lenski2003evolutionary}. This type of model allows for the measurement of selective values \cite{orr2009fitness}, and can be applied to problems larger than living things, such as: the emergence of segregation \cite{schelling1971dynamic}, the transmission of information \cite{zhao2012sihr}, collective movements \cite{vicsek1995novel,reynolds1987flocks} and the evolution of trust \cite{boyd1989mistakes}. Finally, it would be possible to further improve the proposed TAG-game using additional spatial constraints, in analogy with the effect of spatial patterns on sociocultural evolution \cite{danchin2010inclusive,novembre2009spatial}.

The results showed the efficiency of this hybrid model for the functionalization of a decision making (TAG-game) and classification problem (MNIST). Such model could help avoiding the tendency of networks to forget the information learned when they receive new information \cite{mccloskey1989catastrophic}. Furthermore, it could also be consistent with the observation of modules in neuroscience such as: instinctive reactions \cite{rakison2008infants}, some mental disorders (dys \cite{nicolson1993towards}, capgras \cite{anterion2008odd}), memory \cite{mcclelland1995there}, functionality of cerebral lobes \cite{geschwind1979specializations}, and more. One leading result is that the separation of tasks by independent module is more efficient. This suggests it would be more interesting to build a general and flexible model allowing for the emergence of several modules, where the usage of each module is chosen by a 'master' module. This master module would then choose the module to be favored regarding the task, similarly to adversarial networks \cite{goodfellow2014generative}. For instance, in the TAG-game one can imagine a model where the master network would see its intrinsic trophism values, and depending on the opponent's case, choose to use the prey module or the predator one.

\newpage
\supplementarysection
\subsection{Recurrence and virtual inputs}
Recurrence is essential for memory problems, but the gradient tends to vanish when the network is deep \cite{pearlmutter1995gradient}. In our case, the artificial neural network is a feed-forward network, computed from a graph \cite{neubig2017fly}. However, when the adjacency matrix indicates a connection to a downstream layer, the calculation cannot be anticipated. In order to solve this, we separate and store each output in memory. Then, when the network connects to a downstream output, it connects to memory at the $t-1$ time step, which is equivalent to a new input. This recurrence is not standard, and we call it \textit{virtual input pseudo-recurrence}.

\subsection{Hardware and software}
The following hardware and software was used for the computations: Ubuntu 20.04.3, Python 3.8.10, Pytorch 1.7.0 \cite{stevens2020deep}, Numpy 1.18.5 and Scipy 1.5.2 \cite{bressert2012scipy}, Skikit-learn 0.23.2 \cite{pedregosa2011scikit}, Matplotlib 3.3.2 for data visualization \cite{ari2014matplotlib}, Pandas 1.1.3 for data analysis and exploitation\cite{mckinney2011pandas}, Intel Core (i7-4810MQ), NVIDIA with CUDA (Quadro K1100M), Tensorflow Keras 2.3.1 (database) \cite{ketkar2017introduction}, RAM 32GB. Code is available on Github in fabienfrfr/AFF repositories, module named FUNCTIONNAL\_FILLET.

\subsection{Construction of the graph}
The adjacency matrix of each new (initial or random) neural network is randomly constructed. In our model we impose the following constraints: \textit{(i)} layers are fully connected; \textit{(ii)} a connector can connect to one output only (blue nodes in Fig. 3); \textit{(iii)} layers have a minimum of one connector and neuron and \textit{(iv)} at least one connection is made to a back layer. This structure allows for forward graph computation. The maximum numbers of layers is limited to 32 and the numbers of perceptron $N_p$ is defined by a random number between 1 and the number of inputs and outputs. Following such constraints, the minimum number of connector $N_c$ is equal to $N_c = N_p + N_i$ with $N_i$ the number of inputs.

\subsection{Training the ANN}
The following setup is used for training the ANN: One convolutional neural network (CNN) for input \cite{kiranyaz20211d} with $N_{input}$ (input size) group and size one kernel (sparse weight), Linear layer for other \cite{baldi1995learning}, ReLU for hidden layers and input \cite{fukushima1969visual}, Q-Learning (described in Watkins) \cite{watkins1992q}, Loss Function criterion (cross-entropy for classification \cite{gordon2020uses}, smoothL1 for regression and RL problem \cite{pesme2020online}) and optimizer of gradient descent by 'Adam' algorithm \cite{kingma2014adam}.

\subsection{Exploration/Exploitation dilemna}
For the two Markov decision processes, the exploration/exploitation dilemma cannot be directly applied due to the evolution process because we reset the hyperparameters for each generation. Instead, the exploration process is defined by an exponential normalization of the order output of the neuron during training:
\begin{equation}
\label{eq:dilemna}
    P[X_i]= \dfrac{e^{O(X_i)}}{\sum_{i=1}^{N} e^{O(X_i)}}
\end{equation}
with $O$ the order of output response $X_i$. The exponential was chosen to keep the ratio of the primes output separated whatever the number of outputs. The exploitation process consists in selecting the best network when 90\% of the number of episode/batch per generation is reached.

\subsection{TAG-game rules and scoring}
When the agent learns to play both the prey and predator roles, the steps of the TAG-Game are:
\begin{enumerate}
    \item The learning agent is a prey. It must learns to avoid being caught by the predator. The predator always calculates the optimal displacement which minimizes the Euclidean distance between itself and its prey.
    \item When the learning agent is 'caught' by the predator, it then becomes a predator (the roles are permuted) and it must learns to catch the prey. The prey always calculates the optimal displacement which maximizes the Euclidean distance between itself and the predator.
    \item If the learning agent catches the prey, we return to step~1.
\end{enumerate}
In this game, the learning agent has 17 input 'states' ($3\times3$ grid centered and 8 borders), 3 output 'actions' (minimal 3-periodic: $\downarrow$ (Down), $\rightarrow$ (Right), $\nwarrow$ (North-West)) and a neural network in between, where the positional input / output properties on the grid are randomly generated at the start of the experiment. They are then maintained to the offspring. The environment assigns points at each stage of the game and the score is defined as the sum of the points of the agent. The allocation of points to the agent is an important process in reinforcement learning. It needs to be well balanced in order to prevent exploitation from the reinforcement learning, which is likely to exploit flaws in the rules of the game as it tries to optimize the reward. The points are distributed as follows:
\begin{itemize}
    \item -1 for each time step elapsed when the agent is a predator
    \item +10 if the agent catches the prey
    \item +1 for each time step elapsed when the agent is a prey
    \item -10 if the agent is caught by the predator
    \item -5 if the same direction is repeated 3 times
\end{itemize}
These values were estimated using a pre-processing task, which consists of repeated simulations of two random walks in the TAG-game environment: a $12\times12$ grid with periodic boundary conditions. The average number of steps before the two random walks collide for the first time is empirically estimated using 1000 simulations. In each simulation, the starting location of each random walk is assigned randomly with the following constraints: they cannot be identical; they cannot be contiguous with an exception for diagonals, meaning that contiguous locations such as North, South, East, West are forbidden, while North-West, South-East, etc. are allowed. Using these 1000 simulations, the average number of steps before the first collision is estimated to 10. This value is then used for balancing the reward, and setting the values of each point. Therefore, in our grid setup, the learning agent needs to walk at least 10 steps in order to be rewarded with 10 points in a given experience. This allows for distinguishing between a random collision occurrence and a successful strategy. We rely on this empirical estimation rather than an analytical solution as the average number of steps before the first collision of two random walks in two-dimensional lattices with periodic boundary conditions is not trivial.

The last penalty is introduced to promote multi-steps strategies and avoid single-step ones. Multi-steps strategies are sequences of steps aiming to avoid the predator or catch the prey. Single-step strategies are based on one single movement or direction, such as avoiding the predator by moving in diagonal.

\subsection{Phylogenetic Tree}
The phylogenetic tree is the representation of one lineage as an arborescent structure, such as the family tree. Each node is a given individual and each link describes the relationship between individuals. This is illustrated in supplementary Fig. \ref{fig:sup1}. When the lineage becomes super-dominant -when all the $\alpha$ children are always promoted from one generation to the other- then eq. 1 can be approximated as a geometric series:
\begin{equation}
\label{eq:eq2_Lineages}
    \#\text{nodes}_L \approx \sum_{i=1}^{N} \alpha^i = \dfrac{\alpha^{N+1} -1}{\alpha-1}-1
\end{equation}
for large $N$, where $N$ is the number of generations or lifespan of the lineage , which is also the tree's height. The lifespan value of $\#\text{nodes}_L$ of each lineage is then in the following interval:
\begin{equation}
\label{eq:eq3_Lineages}
    \#\text{nodes}_L \in \left[ \alpha (1 + N),\; \dfrac{\alpha^{N+1} -1}{\alpha-1}-1 \right], \quad N > 0
\end{equation}

\begin{figure}
\centering
\includegraphics{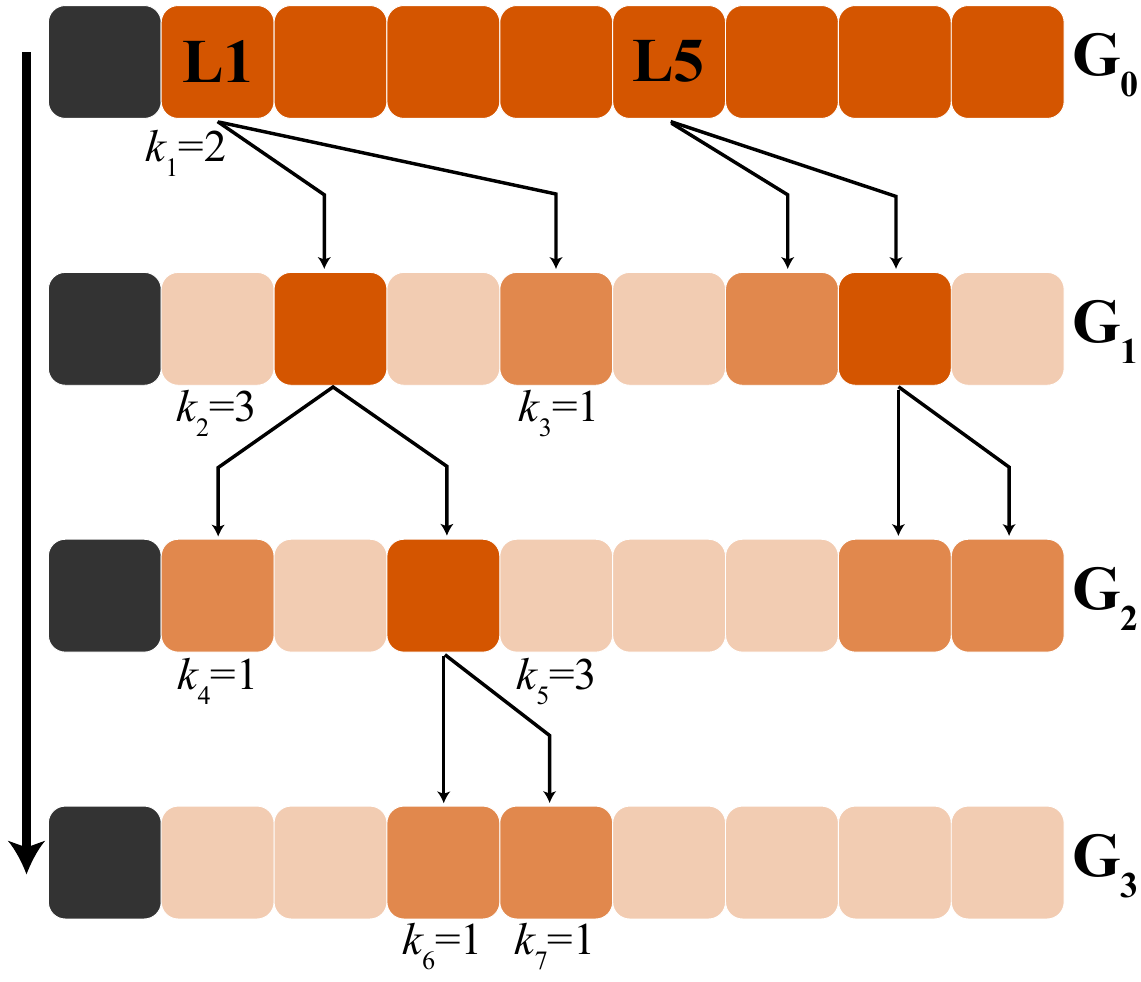}
\caption{The phylogenetic tree. An initial population of 8 networks and 1 control network are generated at the initial generation $G_0$. Two lineages $L1$ and $L5$ are illustrated for $N=3$ generations. The first lineage $L1$ has $\#\text{nodes} = 6$ descendants, while the second lineage $L5$ has $\#\text{nodes} = 4$ descendants (eq. 1).}
\label{fig:sup1}
\end{figure}

\newpage
\bibliographystyle{unsrt}
\bibliography{template}

\end{document}